\documentclass[sn-vancouver,Numbered]{sn-jnl}

\usepackage{graphicx}%
\usepackage{multirow}%
\usepackage{amsmath,amssymb,amsfonts}%
\usepackage{amsthm}%
\usepackage{mathrsfs}%
\usepackage[title]{appendix}%
\usepackage{xcolor}%
\usepackage{textcomp}%
\usepackage{manyfoot}%
\usepackage{booktabs}%
\usepackage{algorithm}%
\usepackage{algorithmicx}%
\usepackage{algpseudocode}%
\usepackage{listings}%

\theoremstyle{thmstyleone}%
\theoremstyle{thmstyletwo}%

\theoremstyle{thmstylethree}%

\begin{document}
\begin{titlepage}

\title[Article Title]{Using Human-like Mechanism to Weaken Effect of Pre-training Weight Bias in Face-Recognition Convolutional Neural Network.}


\author[1]{\fnm{Haojiang} \sur{Ying}}\email{hjying@suda.edu.cn}
\equalcont{These authors contributed equally to this work.}

\author[1]{\fnm{Yi-Fan} \sur{Li}}\email{2013401059@stu.suda.edu.cn}
\equalcont{These authors contributed equally to this work.}

\author*[2]{\fnm{Yiyang} \sur{Chen}}\email{yychen90@suda.edu.cn}

\affil[1]{\orgdiv{Department of psychology}, \orgname{Soochow university}, \orgaddress{\street{No.1, Shizi Street}, \city{City}, \postcode{215006}, \state{Jiangsu}, \country{China}}}

\affil*[2]{\orgdiv{School of Mechanical and Electric Engineering}, \orgname{Soochow university}, \orgaddress{\street{No. 8, Jixue Road}, \city{Suzhou}, \postcode{215137}, \state{Jiangsu}, \country{China}}}

\maketitle

\textbf{Author Contributions}\\
\\
\textbf{H. Ying}: Conceptualization, Data Analysis, Funding \& Resources Acquisition, and Writing (revision and final draft). \textbf{Y.-F. Li}: Coding, Data Analysis, Data Acquisition, and Writing (initial draft and suggestion). \textbf{Y. Chen}: Conceptualization, Funding \& Resources Acquisition, and Writing (suggestion and final draft).

\end{titlepage}



\section*{Abstract}

Convolutional neural network (CNN), as an important model in artificial intelligence, has been widely used and studied in different disciplines. The computational mechanisms of CNNs are still not fully revealed due to the their complex nature. In this study, we focused on 4 extensively studied CNNs (AlexNet, VGG11, VGG13, \& VGG16) which has been analyzed as human-like models by neuroscientists with ample evidence. We trained these CNNs to emotion valence classification task by transfer learning. Comparing their performance with human data, the data unveiled that these CNNs would partly perform as human does. We then update the object-based AlexNet using self-attention mechanism based on neuroscience and behavioral data. The updated FE-AlexNet outperformed all the other tested CNNs and closely resembles human perception. The results further unveil the computational mechanisms of these CNNs. Moreover, this study offers a new paradigm to better understand and improve CNN performance via human data.\\
\\
\textbf{keywords:} Convolutional neural network, Face perception, Psychophysics, Transfer learning, Computer vision.

\section{Introduction}\label{sec1}

Convolutional neural network (CNN) is an important model in artificial intelligence. It was first inspired by Hubel and Weisiel's works \cite{hubel1959receptive,hubel1962receptive} on the visual nervous system of cats, which processes visual input signals by layered convolution to achieve visual tasks. CNNs have been widely used in tremendous research areas, such as  customer Speech analysis \cite{murugaiyan2023aspect}, biomedical image segmentation \cite{iqbal2022ff}, radar image recognition \cite{yue2021novel} etc. In fact, in many tasks, CNNs have equaled or surpassed human performance \cite{weyand2016planet,yu2017sketch,larmuseau2021race}. Despite the appealing performance of CNNs, their internal mechanisms are still not fully revealed due to their complex nonlinear computational processes and millions of parameters. In other words, neural networks can be considered as a `black box', and their principles of operation and interpretability still need to be further explored.

In the previous decades, with the substantial development of computer processing power, CNN has been systematically improved in advanced structure and achieves a better performance. Based on the original LeNet \cite{lecun1998gradient}, researchers have proposed state-of-the-art models such as AlexNet \cite{krizhevsky2012imagenet}, VGGNet \cite{simonyan2014very}, ResNet \cite{he2016deep}, SENet \cite{hu2018squeeze}, EfficientNet \cite{tan2019efficientnet}, etc., by deepening the network, adjusting the perceptual field size, introducing multiple nonlinear activation functions, improving the convolution process, etc. These networks usually consist of tens to hundreds of layers of feature extraction computational layers, which are also commonly referred to as "deep" convolutional neural networks (DCNN). The structure of convolutional neural networks is inspired by the biological visual system, and thus in some recent studies, some neuroscientists and cognitive scientists have tried to further establish the connection between CNNs and human cognitive and neural systems. They study CNN as a brain-like model, such as using neural networks as a model for the human visual system \cite{lindsay2021convolutional,luo2016understanding}, or using them to solve some problems that cannot be directly experimented on the human nervous system \cite{barrett2019analyzing}. In addition, some neuroscience findings have also been applied to CNNs as inspiration, such as spiking neural networks (SNN) \cite{ghosh2009spiking,tavanaei2019deep,yang2022spiking,rathi2023exploring} based on the leaky integrate and fired model \cite{gerstner2014neuronal}, PredNet \cite{lotter2017deep} based on predictive coding theory \cite{rao1999predictive}, and solving catastrophic forgetting problems of neural networks using brain-inspired replay \cite{van2020brain}, etc.

Therefore, testing the performance of CNNs using cognitive science approaches is very feasible to improve their interpretability, e.g., using the experimental paradigm of reverse correlation to reconstruct and compare the prototype of CNN with that of humans \cite{song2021implementation}, as well as using the psychophysical method to compare the performance of CNN with that of humans \cite{zhou2022emerged,li2022disrupted}. Transfer learning \cite{bengio2012deep} involves using a pre-trained model to initialize the weights of a new model, and then fine-tuning the model by training it on a new dataset for a specific task. This approach can effectively use the features learned by the pre-trained model on large datasets, thus improving the performance and training speed of the new model. Transfer learning has emerged as a new framework of machine learning, which is based on deep learning \cite{zhu2021phase}. 

In this study, we trained four classic CNNs that have been extensively studied-AlexNet, VGG11, VGG13, and VGG16-for the face emotion valance task based on pre-trained model weights of objects and faces for transfer learning, and then tested them on a set of isometrically varying face emotion valance test sets. We first analyzed the attentional tendency of the feature extraction layer of CNN using the LayerCAM \cite{jiang2021layercam} method, and then masked important information in human faces (eyes, nose, mouth) using the experimental paradigm of inverse correlation. We conducted behavioral experiments and performed similar tasks on humans, while plotting the psychophysical function (a variance of the cumulative Gaussian function) of humans and CNNs, and their points of subjective equality (PSE) \cite{webster2004adaptation} were compared. Also finally we proposed an improved method of AlexNet, enhancing its performance to have a more human-like performance.

\section{Methods}

\begin{figure*}[htbp]
    \centering
    \includegraphics[width=0.75\textwidth]{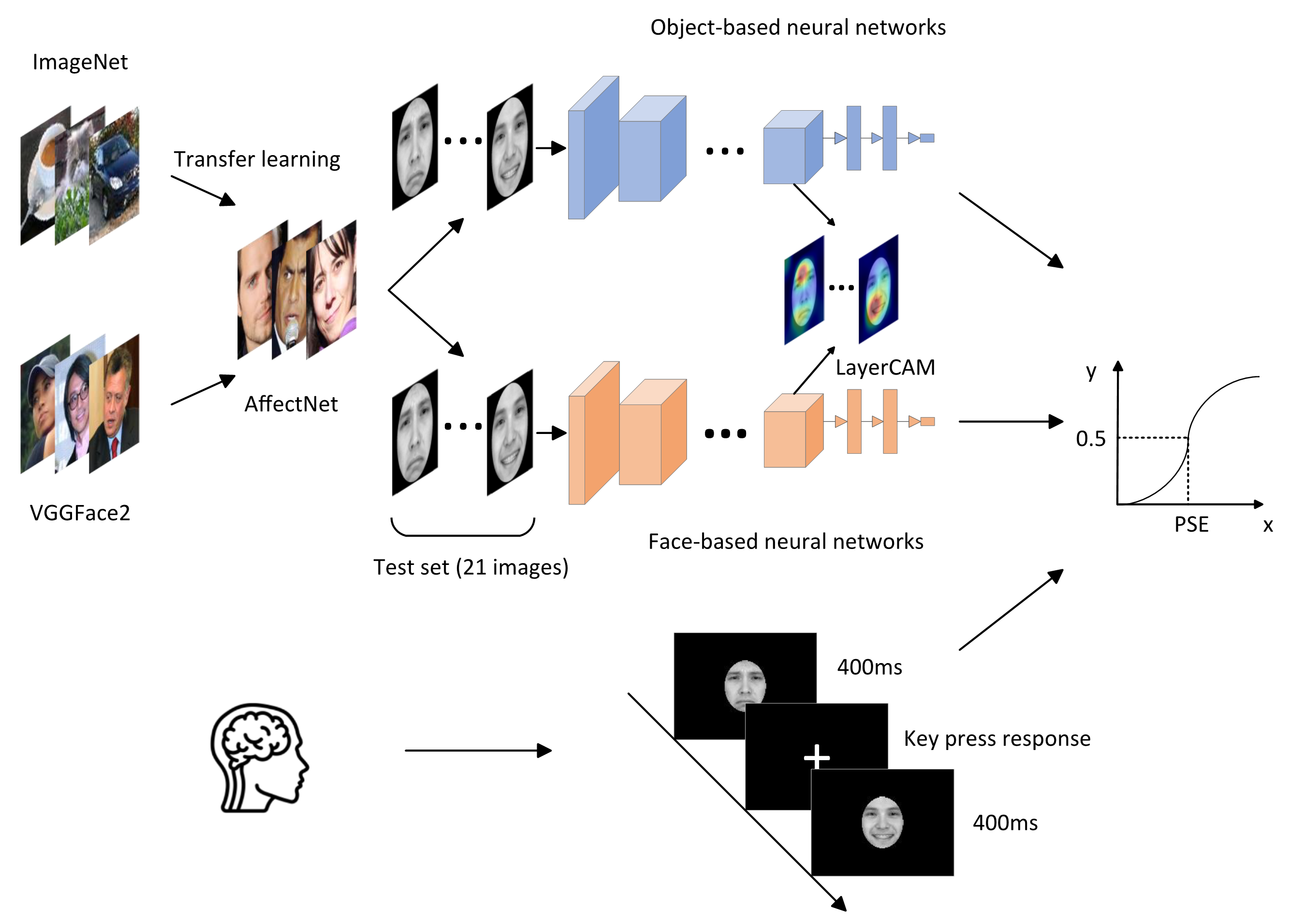}
    \caption{The overall experimental procedure.  The ImageNet pre-training weights were provided by Pytorch, and the VGGFace2 pre-training weights were pre-trained using our filtered 1000-class VGGFace2 dataset. Both pre-trained models were used to perform the emotion valence classification task by transfer learning on the AffectNet dataset. In Experiments 1 and 2, LayeCAM was generated for the last convolutional layer of the neural network, and for humans we designed a simple behavioral experiment in which subjects were asked to perform a binary forced choice of emotion valence. We used the predictions from the neural network model and the data from the human behavioral experiment to plot psychometric functions and to statistically measure their PSE, which was done in all three experiments. Brain icon made by Smashicons from www.flaticon.com.}
    \label{procedure}
\end{figure*}

This study consists of three experiments. The schematic of the experimental procedure is shown in Figure \ref{procedure}. First, we tested human performance (psychometric functions) and used it as a baseline to test DCNN performance. In Experiment 1, we performed the same task as humans for four DCNNs (AlexNet, VGG11, VGG13, VGG16) trained with object-based pre-training (ImageNet) and face-based pre-training weights (VGGFace2), i.e., judging emotion valence. The similarities and differences between their PSEs and those of humans will be contrasted. In this experiment, we also used the LayerCAM method to visualize the last layer of the neural network feature extraction, in order to observe the differences in the attentional areas of different pre-trained and different DCNNs. In Experiment 2, we applied a reverse correlation experimental paradigm, where important components of face information (eyes, nose, mouth) were masked and presented as inputs to the DCNNs for prediction, again comparing their LayerCAMs and PSEs. In Experiment 3, our new and improved method is applied to AlexNet to perform the tasks in the above two experiments and compare their psychometric functions and PSEs.

\subsection{Artificial Neural Networks}

This section introduces the convolutional neural networks we are studying, the training environment, and the selection and processing of the dataset.

\subsubsection{Neural network Architectures}

In this study, we employed transfer learning deep convolutional neural networks (CNNs) as a facial emotion recognition system. Specifically, we utilized four CNN models, including the popular AlexNet and the VGGNets with varying depths. The AlexNet comprises eight layers, with five convolutional layers and three fully connected (FC) layers. The convolutional layers extract features from the images, and the FC layers perform the final weighted classification. To prevent gradient explosion or disappearance, it is applied the Rectified Linear Unit (ReLU) 
 \cite{glorot2011deep} non-linear activation function after each layer (except the last one). It also includes MaxPool operations after the convolutional layers of 1, 2, and 5 layers to compress features and alleviate over-sensitivity of the convolutional layers to position. Additionally, it is utilized the Dropout \cite{srivastava2014dropout} operation with \textit{p} = 0.5 before the first two FC layers to discard some neural links and minimize overfitting.

VGGNets \cite{simonyan2014very} are deeper than AlexNet, with smaller convolutional kernel operation and VGG blocks structure. The typical VGGNet consists of five VGG blocks and three FC layers, similar to AlexNet. We used VGG11, VGG13, and VGG16, which represent the 11-layer, 13-layer, and 16-layer VGG networks \cite{simonyan2014very}, respectively. they were applied ReLU non-linear activation after each layer except the last one, MaxPool operation after each VGG block, and Dropout method in the first two FC layers.

Finally, we construct feature excitation (FE) AlexNet based on AlexNet, in the feature extraction layers (convolutional layers) is the same as AlexNet, we modified the classification layers (fully connected layers), the modification details refer to Figure \ref{fealex}.

We implemented and developed our neural networks using the PyTorch \cite{paszke2019pytorch} deep learning framework, an open-source platform with a default kaiming\_normal method for initialization. 

\subsubsection{Device and environment}
The neural network framework was built using PyTorch, and the code was implemented in Anaconda environment with Python environment. Pre-trained models are trained in the Beijing Super Cloud Computing Centre (an online computing platform). Our computing environment on this platform is PyTorch version 1.12.1, compute unified device architecture (CUDA) version 11.3, using an A100 GPU. Transfer learning of all networks and image pre-processing were performed on a computer running Windows 10 (Microsoft, WA) OS with Intel Core i9 9900T CPU, 32GB RAM, and an RTX2070 graphics processing unit (GPU). The PyTorch version is 1.12.1, and CUDA version is 11.6.

\subsubsection{Dataset processing}
For the pre-training, face-based training we chose the VGGFace2 dataset \cite{cao2018vggface2}. To reduce the computational pressure and the number of classifications corresponding to ImageNet pre-training, we randomly selected 1,000 subjects and processed them in the same way in the training and validation sets. This resulted in a training set of 291,502 images with 1,000 classes and a validation set of 72,378 images with 1,000 classes

For transfer learning, we used AffectNet dataset \cite{mollahosseini2017affectnet}. AffectNet is a large facial expression dataset with around 0.4 million images manually labeled for the presence of eight (neutral, happy, angry, sad, fear, surprise, disgust, contempt) facial expressions along with the intensity of valence and arousal. Floating point values for valence are provided in the dataset and we define values less than 0 as `negative' and values greater than or equal to 0 as `positive'. Similarly, we performed a random sampling of the dataset. The above operation was applied to both the training and validation sets, and we ended up with a training set of 28,765 images for binary classification and a validation set of 2,999 images for binary classification.

\subsubsection{Neural network training}
For the neural networks we performed two phases of training. At the first phase (the pre-training phase), for the object-based weights we directly used the pre-training weights provided by PyTorch trained on the ImageNet dataset and for the face-based weights we used our custom 1,000 classes VGGFace2 dataset for training. For the second phase - the transfer learning phase - we used ImageNet weights and our own trained VGGFace2 weights as the initialization parameters for the model, respectively. For the transfer learning approach we chose to freeze the feature extraction layer of the networks and keep the fully connected layer of the networks, the classification layer, for further training.

Based on our past experimental experience, we used the same settings in all training, set batchsize to 32 and learning rate to $5\times 10^{-5}$ and saved the model parameter data with the highest accuracy on the validation set as the model for the following experiments. The network was trained with 60 epochs in pre-training and 20 epochs in transfer learning. Before the actual training, we also performed random clipping of the training set and random horizontal flipping to enhance our dataset.

\subsection{Behavior Experiment}

The human data were from an existing experiment testing the same perceptual task. We reused the data as a benchmark to test the performance of the DCNNs.

\subsubsection{Participant Information}

$50$ human participants (mean age = $21.2$; $37$ females and $13$ males) volunteered for the ethnicity perception experiment (part of data from a study currently under review). They offered written consent before the experiment. This study was approved by the Ethics Committee of Soochow University. 

\subsubsection{Experiment procedure}

Here, we briefly introduce its procedure. The procedure of this experiment is adapted from a recent cognitive study using psychophysics methods \cite{ying2020temporal}. 

In general, participants were asked to judge the perceived ethnicity of the testing faces (one at a time) via a Two-Alternative-Forced-Choice (happy or sad) paradigm by pressing A (happy) or S (sad) on the keyboard. Every participant completed $210$ trials ($7$ testing faces $\times 2=39$ repetitions) with randomized orders.

The testing faces are 21 faces generated in a similar way as \cite{ying2020temporal}. We first generated 21 faces with different levels of emotional expression from the same averaged identity (of all female faces from the Taiwanese Facial Expression Image Database (TFEID; \cite{chen2007taiwanese}).), morphed from sad(coded as 0\% of happiness) to neutral (coded as50\% of happiness) to happy(coded as 100\% of happiness) in 5\% steps (via Webmorph software.). We selected all 21 target faces for CNN test and 7 target faces for behavior experiment, with 0\%, 20\%, 30\%, 50\%, 70\%, 80\%, and 100\% of happiness.

The human behavioral experiment was conducted on a PC running Matlab R2016a (MathWorks) via PsychToolBox extensions \cite{brainard1997psychophysics,pelli1997videotoolbox} with a 27-inch LCD monitor (spatial resolution $1920 \times 1080$ pixels, refresh rate $120$ Hz. During the experiment, participants sat in an adjustable chair, with their chins resting on a chin rest which was placed at $85cm$ away from the monitor, and each pixel subtended $0.025$ on the screen.

\subsubsection{Comparison Using Psychophysics Methods}

The results of the psychophysical experiment of human participants were used as the baseline to evaluate the performance of the neural network at other variable levels. Psychophysical experimental methods were adapted from previous psychological studies \cite{webster2004adaptation} that asked each subject to judge the emotional valence classification of the face tested from two options (press A (Happy) or S (Sad) on the keyboard to make a judgment from two options.). Each subject performed 210 trials, (7 faces × 30 repetitions), in a randomized order. For the machine, the same 21 face (the 7 faces tested on human participants were selected from these 21 faces) images were given to the machine for recognition, and after a complex CNN operation, a Softmax function was used for the resultant values to calculate their recognition probability for the resulting faces. Based on previous studies in \cite{webster2004adaptation}, different people with different standards of ethnicity judgments shall have different points of subjective equality (PSE). The PSE of each curve was then measured. With the above data, we can derive the PSE for humans and CNNs, and we can then compare the PSE of the two types of curves. We used the \textit{t}-test to compare the performance of each CNN against the PSEs of $50$ participants (the PSE of each CNN against the mean and deviation of the human participants). Based on this, we will also submit the masked face images to CNNs for the same operation and also baseline with the above human data to go further with our experimental study.

\subsection{Attention Area Analysis of Neural Networks and Human}
Previous studies \cite{song2021implementation,li2022disrupted,zhou2022emerged,zhang2018unreasonable} mentioned all point to the conclusion that there are similarities between humans and neural networks in information acquisition and processing. To validate our hypothesis about similarities and differences in information processing between neural networks and humans, we used different neural network interpretability methods. The methods can be divided into four types: visual, interference-based, knowledge-based, and causal interpretation methods. We chose to compare neural networks and humans from the perspective of attention, using eye-tracking methods for humans and class activation maps (CAM) \cite{zhou2016learning} for neural networks. Grad-CAM \cite{selvaraju2017grad} is a widely used method that solves some of the shortcomings of CAM by back-propagating the gradient and using the ReLU activation function. However, few studies have combined gradient-based CAM methods with eye-tracking analysis. Our paper aims to conduct experimental exploration of this relatively empty field using the newer LayerCAM \cite{jiang2021layercam} method proposed by Jiang et al. based on a modified version of Grad-CAM, which offers a more refined performance than Grad-CAM. LayerCAM can be expressed in the formula as
\begin{equation}
M_{(x,y)}^{c} =\text{ReLU}( \sum_{k}(\text{ReLU}\frac{\partial y^c}{\partial A_{xy}^{k}}  )A_{xy}^{k})
\end{equation}
where $\frac{\partial y^c}{\partial A_{xy}^{k}}$  represents the gradient of the corresponding class on the pixel, and then use ReLU to get a positive value to get the pixel-level weight, and then use this weight to make a linear combination with the original pixel $A_{xy}^{k}$, and finally the combination between channels, and then after ReLU processing to get LayerCAM $M_{(x,y)}^{c}$.

\subsection{Feature Excitation (FE) AlexNet}

Here we base on the SE block proposed in SENet \cite{hu2018squeeze}, and considering that the block mainly consists of two fully connected layers, we use this block to replace the first two fully connected layers of AlexNet, and since it corresponds to ‘Squeeze and Excitation’ computation only after convolutional feature extraction, and we use only excitation and scale operations. Since the implementation area is different from most applications, we make some modifications to the SE block, in particular, the SE block converts multidimensional convolutional features into one-dimensional feature vectors by global average pooling to make them suitable for the fully connected layer, where the number of output connections decreases exponentially. In our implementation, the final output of the convolutional computation has been one-dimensionally flattened using nn.flatten, so instead of using global average pooling, we directly use the one-dimensionally flattened feature vector for full connectivity. The second fully connected layer expands the number of connections to the most original input. Here, we set the compression ratio to 4:1, so that we do not exploit the squeeze, but rather its excitation part. A sigmoid function is used at the end of the excitation operation to form a feature importance weight matrix, which is used to scale the original feature vector (i.e., multiply the corresponding elements). Note that no bias is introduced in this part of the fully connected layer, and the final output obtained by such an operation is then passed through a fully connected layer to output the binary tensor and finally calculate the probability of each outcome by softmax. FE layer can be expressed in the formula as 

\begin{equation}
X_{i}^{'}  = X_{i}\times \text{Sigmoid}(W_{2}(\text{ReLU}(W_{1}\times X_{i})))
\end{equation}
where $X_{i}$ represents a single feature element and $X_{i}^{'}$ represents the feature element after excitation. $W_{1} \in R^{9216 \times 2304}$ represents the weights of the first unbiased fully connected layer and $W_{2} \in R^{2304 \times 9216}$ represents the weights of the second unbiased fully connected layer. All elements are combined into a feature vector to participate in the final fully connected classification computation.

\begin{table}[htbp]
\centering
\caption{Table I represents the PSE statistics of the prediction equations for the four classical neural networks and the modified FE-AlexNet using isometric varying emotional valence face image inputs under two training strategies, where the experimental human behavioral data is used as the baseline and the p-values are are Bonferroni corrected.}
\begin{tabular}{lllllll}
\hline
                &         &  &  PSE   & \textit{t}(49)   & \textit{p} \\ \hline
Object-based    & AlexNet &  &  $\blacktriangle$ & $\blacktriangle$    &  $\blacktriangle$    \\
                & VGG11   &  &  0.701 & 9.320            &  $<$.001     \\
                & VGG13   &  &  0.494 & -2.116           &  $<$.001     \\
                & VGG16   &  &  0.703 & 9.436            &  $<$.001    \\
                & \textbf{FE-Alexnet}   &  &  \textbf{0.501} & \textbf{-1.691}      &  \textbf{0.875}       \\\hline
Face-based      & AlexNet &  &  0.670 & 7.630            &  $<$.001     \\
                & VGG11   &  &  0.714 & 10.072           &  $<$.001     \\
                & VGG13   &  &  0.715 & 10.116           &  $<$.001     \\
                & VGG16   &  &  0.622 & 4.989            &  $<$.001     \\
                & \textbf{FE-Alexnet}   &  &  \textbf{0.663} & \textbf{7.232}       &  \textbf{$<$.001}    \\\hline
\label{tab1}
\end{tabular}
    \begin{tablenotes}
        \footnotesize
        \item $\blacktriangle$ Indicates that the data had a large bias. thus the results are not meaningful and had been discarded for analysis.
    \end{tablenotes}
\end{table}

\section{Results}\label{sec2}

There are three experiments in this study. In the first experiment, we tested the neural network using psychometric functions and PSE \cite{webster2004adaptation} and LayerCAM \cite{jiang2021layercam}. In the second experiment, we applied a previously used reverse correlation experimental paradigm \cite{li2022disrupted,song2021implementation}. In this sense, we replaced the test images with ones that obscure important information about the faces, thus we can reversely find the key image segments that determines an accurate processing of the image. In the third experiment, we introduced a new CNN, FE-AlexNet, performing both a general emotion valence classification task and an emotion valence classification task with lossy inputs, and fitting its psychometric function and generated the point of subjective equality (PSE). All of the above experiments used human behavioral experimental data as a baseline.

\subsection{The Experiment 1: Comparing performance with object-based and face-based transfer learning}

First, in this experiment, we generated and plotted LayerCAM \cite{jiang2021layercam} attention hot map for CNNs pre-trained based on different tasks (objects and faces). Then we simultaneously plotted psychophysical curves based on human experimental data derived from behavioral experiments and test set data from CNNs, and counted their PSEs. 

\subsubsection{Results of Experiment 1}

\begin{figure*}[t!]
    \centering
    \includegraphics[width=1\textwidth]{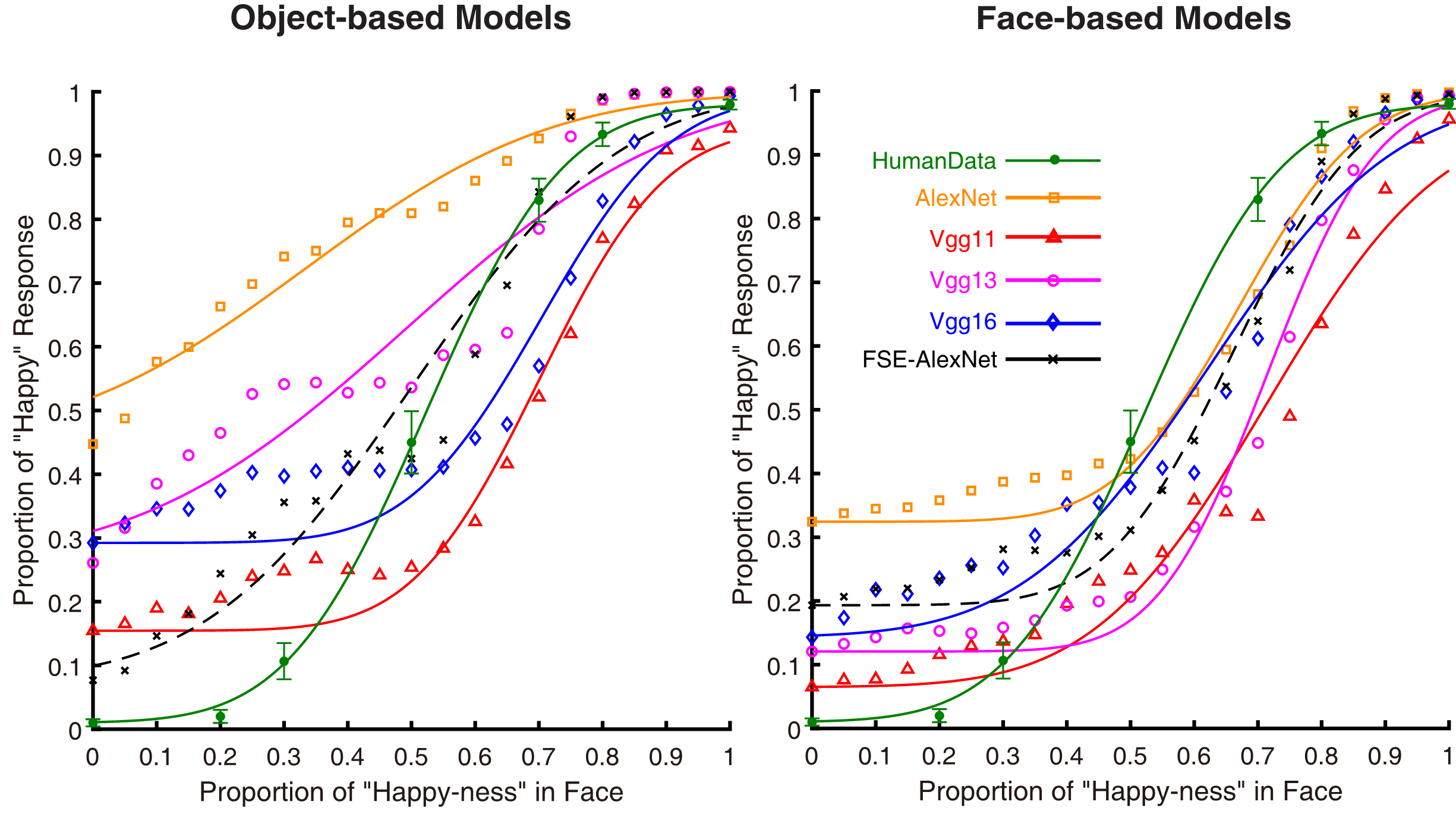}
    \caption{Psychometric functions for Experiment 1. Presented here are the psychometric functions for four different levels of the network and the modified FE-AlexNet under two training strategies as well as for humans. Green lines and symbols represent human participants, yellow lines and symbols represent AlexNet, red lines and symbols represent VGG11, magenta lines and symbols represent VGG13, blue lines and symbols represent VGG16, and black dashed lines represent FE-AlexNet. The horizontal coordinate is the rate of change of the 21 face images with uniform changes in emotional validity in steps of five percent, and the vertical coordinate is the prediction probability of the neural network. For the human subjects' data, the vertical coordinate is the accuracy of identifying each face emotion image.}
    \label{psf1}
\end{figure*}

\begin{figure*}[htbp]
    \centering
    \includegraphics[width=1\textwidth]{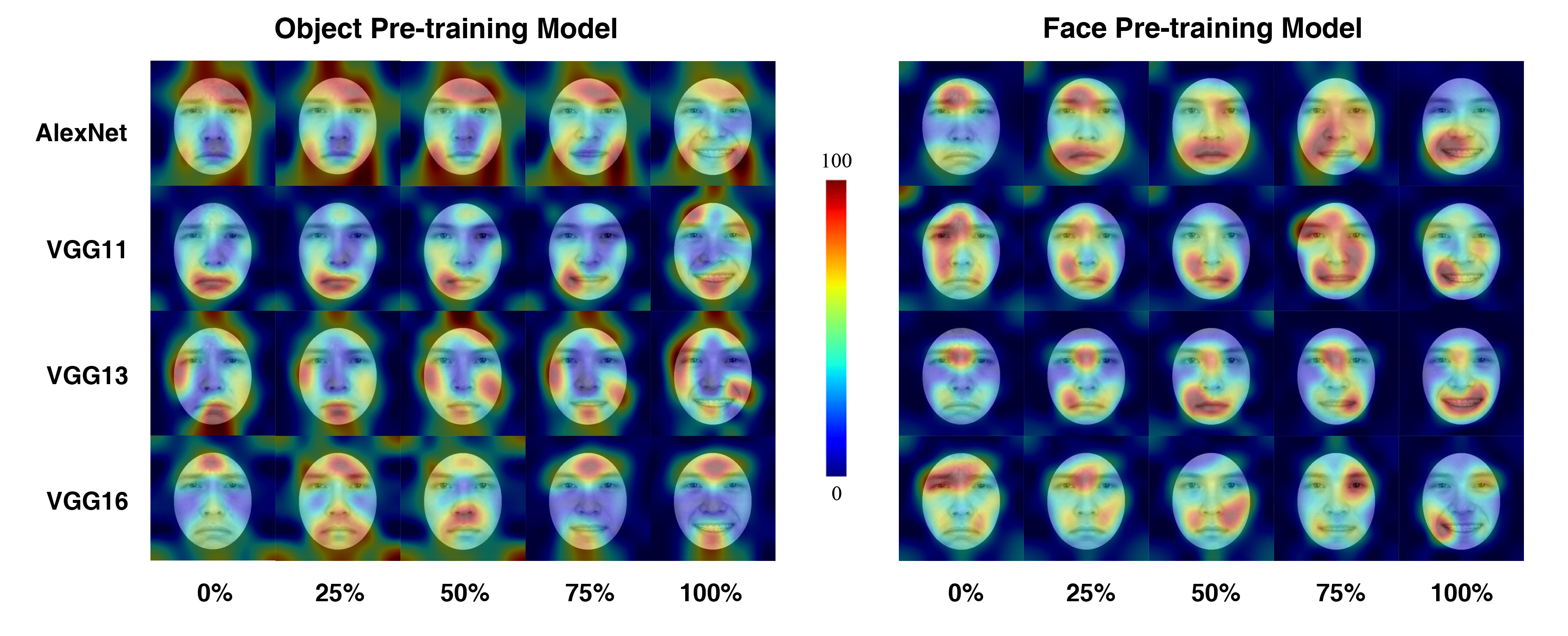}
    \caption{LayerCAM output of Experiment 1. The activation output of the last layer of the convolutional layer for four different layers of the network (AlexNet, VGG11, VGG13, VGG16) with two training strategies (object pre-training and face pre-training) is presented here. Here the activation map output is presented for each network in steps of 25\% from the lowest to the highest emotion valence.}
    \label{exp1}
\end{figure*}

The psychometric functions are shown in Figure \ref{psf1}. Experimental results from human participants (green lines) fit the psychometric function (M = 0.532, SEM = 0.018) well obviously. Psychometric functions for face-based and object-based pre-trained neural networks are shown in Figure \ref{psf1}A 
 and Figure \ref{psf1}B. The two training strategies show a clear difference in performance as the object-based training neural networks exhibit more dispersion and obtain a larger bias in the fitting results compared to the face-based neural networks, especially, AlexNet and VGG13. The \textit{t}-test results of PSE (point of subjective equality) are shown in Table \ref{tab1} (the \textit{p}-values are Bonferroni corrected and we define data for which the y-value of the psychometric fit curve does not intersect with 0.5 as meaningless and is discarded in the analysis, as indicated in the table, and this is also performed in the following experiment). Although the PSE results for most networks (except for AlexNet, which produces pointless results) still differ significantly from those of humans (\textit{p} $<$ 0.001), for networks object and face pre-training still yield large differences, with AlexNet (object, meaningless and face, \textit{t}(49) = 0.670) and VGG13 (object, \textit{t}(49) = 0.494 and face, \textit{t}(49) = 0.715) yielding more significant differences compared to VGG11 (object, \textit{t}(49) = 0.701 and face, \textit{t}(49) = 0.714) and VGG16 (object, \textit{t}(49) = 0.703 and face, \textit{t}(49) = 0.622). The experimental results of the class activation map are shown in Figure \ref{exp1}. It is intuitive to show the differences in the attention patterns of the neural networks under the two training strategies. In the object-trained neural network, with the exception of VGG11, the other networks focus more on non-face information, while in the face-trained neural network the key face information is well captured, and there is a tendency for this information to focus more on the mouth as the facial emotion valence increases. 

\subsubsection{Brief Discussion of Experiment 1}

In general, the results suggest that different prior knowledge (i.e., pre-train weight) influences the attentional tendencies of neural networks. 

In this experiment, we measured the outcome of the transfer learning by fitting psychometric functions and LayerCAM to different neural networks. These results indicate that the CNNs with different prior knowledge (i.e., pre-train weight) have significantly different performance: worse performance for different-task transfer (object to face) than for similar-task transfer (face to face). This result is reasonable. Previous studies on transfer learning in machine learning and studies on learning strategies in cognitive science have consensus that different prior knowledge affects the cognitive patterns of neural systems. 

Appropriate knowledge reuse can help to improve the effectiveness and efficiency of training; however, a mixture of widely varying probability distributions may have a negative impact on transfer learning \cite{bali2019multifactorial,wu2021online, zhang2022survey}, and this theory also applies in higher-level cognitive systems \cite{goyal2022inductive}. In addition, the experimental results show that the attentional tendency of the face-trained CNNs tend to focus more on the mouth as the facial emotion valence increases, which is also consistent with previous eye-tracking studies on humans \cite{schurgin2014eye}. Therefore, the face-trained CNNs not only have similar physical resemblance to human neural system, but also have computational resemblance to human perception. Consequently, it is possible to use human neuroscience and cognitive science data to improve the structure and the performance of the CNNs.

\begin{table}[t!]
\centering
\caption{Table II represents the PSE statistics of the prediction equations for the four classical neural networks and the modified FE-AlexNet under object pre-training using an isometric variation of the key face information with emotionally effective face image input masked, where the experimental human behavioral data is used as the baseline and the p-values are Bonferroni corrected.}
\begin{tabular}{lllllll}
\hline
                &         &  &  PSE   & \textit{t}(49)   & \textit{p}  \\ \hline
Eyes            & AlexNet &  &  $\blacktriangle$ & $\blacktriangle$            &  $\blacktriangle$      \\
                & VGG11   &  &  0.730 & 10.923           &  $<$.001    \\
                & VGG13   &  &  0.640 & 5.989            &  $<$.001    \\
                & VGG16   &  &  0.712 & 9.923            &  $<$.001    \\\
                & \textbf{FE-AlexNet}&  &  $\blacktriangle$ & $\blacktriangle$        &  $\blacktriangle$   \\\hline
Nose            & AlexNet &  &  $\blacktriangle$ & $\blacktriangle$          &  $\blacktriangle$    \\
                & VGG11   &  &  0.605 & 4.050            &  0.001      \\
                & VGG13   &  &  $\blacktriangle$ & $\blacktriangle$           &  $\blacktriangle$      \\
                & VGG16   &  &  $\blacktriangle$ & $\blacktriangle$           &  $\blacktriangle$      \\
                & \textbf{FE-AlexNet} &  &  \textbf{0.186} & \textbf{-19.138}       &  \textbf{$<$.001}     \\\hline
Mouth           & AlexNet &  &  $\blacktriangle$ & $\blacktriangle$           &  $\blacktriangle$      \\
                & VGG11   &  &  0.641 & 6.050            &  $<$.001    \\
                & VGG13   &  &  $\blacktriangle$ & $\blacktriangle$            &  $\blacktriangle$    \\
                & VGG16   &  &  $\blacktriangle$ & $\blacktriangle$           &  $\blacktriangle$    \\
                & \textbf{FE-AlexNet} &  &  \textbf{0.532} & \textbf{-0.011}        &  \textbf{1.000}      \\\hline
\label{tab2}
\end{tabular}
    \begin{tablenotes}
        \footnotesize
        \item $\blacktriangle$ Indicates that the data had a large bias. thus the results are not meaningful and had been discarded for analysis.
    \end{tablenotes}
\end{table}

\subsection{The Experiment 2: Testing CNNs with Essential Facial Features Unavailable}

The first experiment suggests that the CNNs trained based on different tasks seem to have different information processing mechanisms on the very same task. To further verify this difference and to propose an explanation for this phenomenon, in this experiment, we applied the inverse correlation experimental paradigm, suggested by \cite{song2021implementation,li2022disrupted}, by masking important essential information in faces (eyes, nose, mouth) and used these disrupt visual input to CNNs. The experimental manipulation is the same as in Experiment 1.

\subsubsection{Results of Experiment 2}

\begin{figure*}[t!]
    \centering
    \includegraphics[width=1\textwidth]{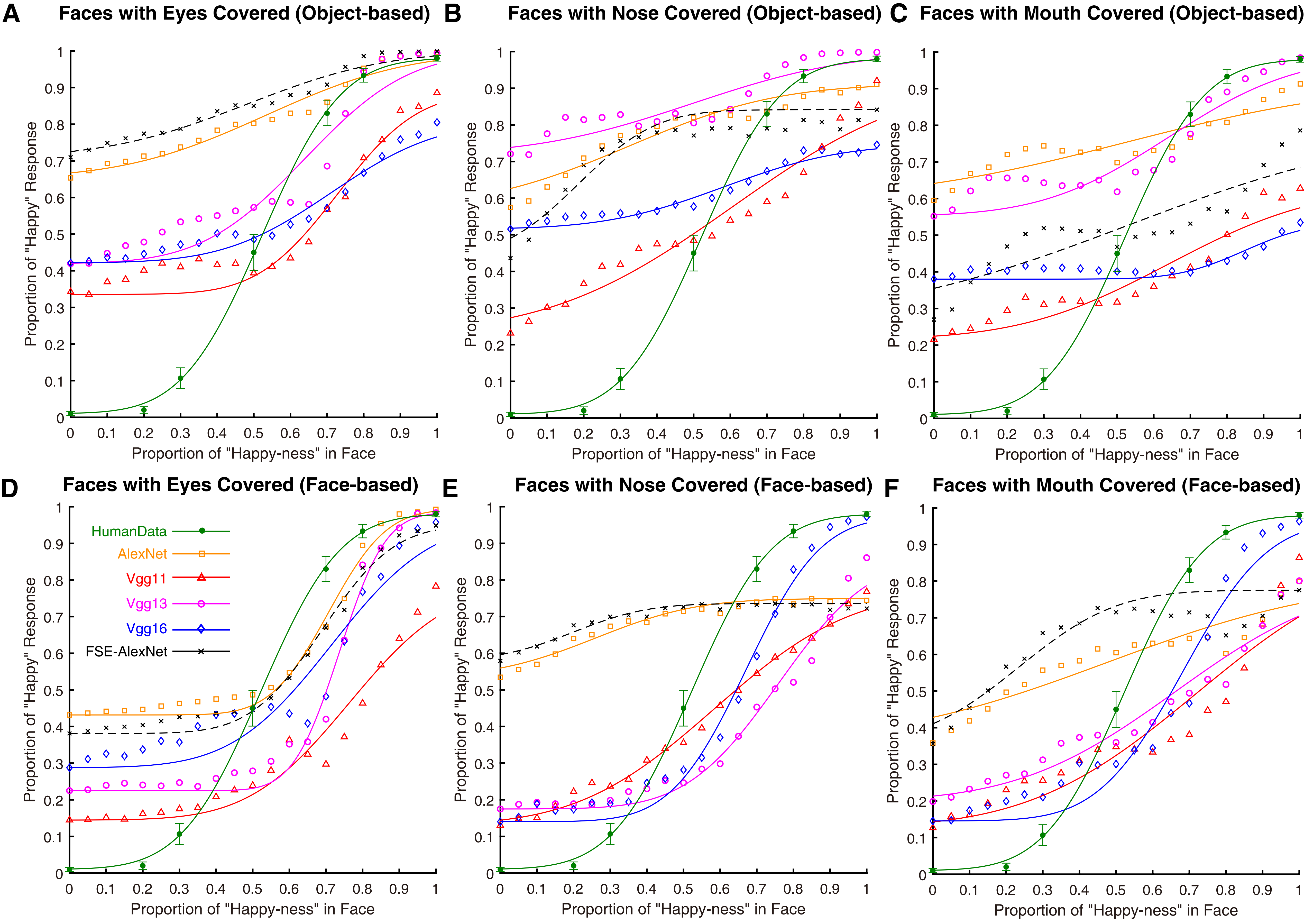}
    \caption{Psychometric functions for Experiment 2. Presented here are the psychometric functions for four different levels of the network and the modified FE-AlexNet under two training strategies as well as for humans. Green lines and symbols represent human participants, yellow lines and symbols represent AlexNet, red lines and symbols represent VGG11, magenta lines and symbols represent VGG13, blue lines and symbols represent VGG16, and black dashed lines represent FE-AlexNet.  The horizontal coordinate is the rate of change of the 21 face images with uniform changes in the emotional valence of the key face information (eyes, nose, mouth) obscured in steps of five percent, and the vertical coordinate is the prediction probability of the neural network. The data for human subjects are the same as for Experiment 1.}
    \label{psf2}
\end{figure*}

\begin{figure*}[t!]
    \centering
    \includegraphics[width=1\textwidth]{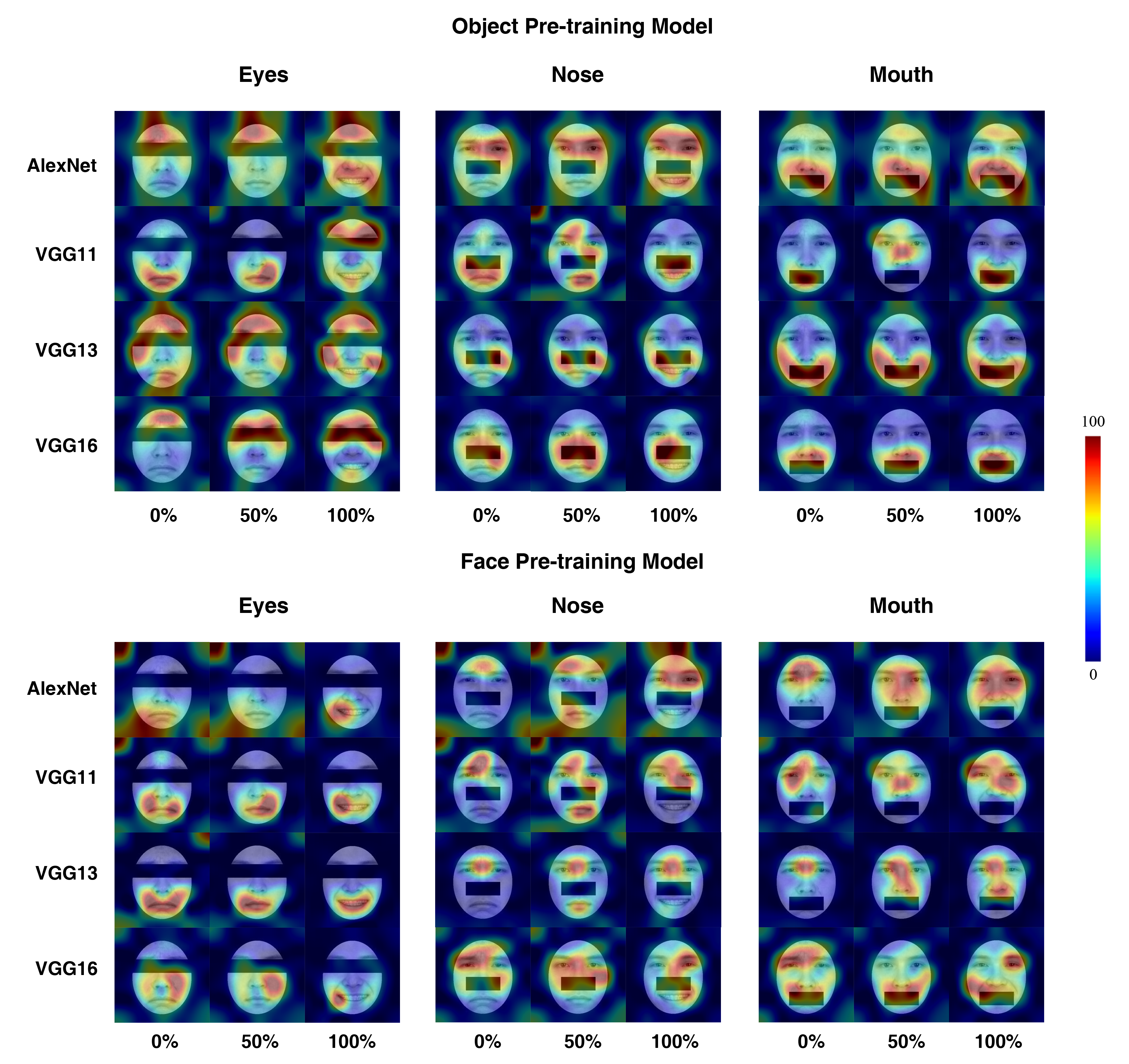}
    \caption{The LayerCAM output of Experiment 2. The activation map output is also presented here for the last convolutional layer of the four different layers of the network under the two training strategies, where the input images are the lowest, neutral, and highest valence emotional images of the face with the key information of the face (eyes, nose, mouth) masked respectively. The computed output of the activation map is presented here.}
    \label{exp2}
\end{figure*}

The psychometric functions for Experiment 2 are plotted in Figure \ref{psf2}, and the PSEs are shown in Table \ref{tab2}, Table \ref{tab3} (the \textit{p}-values are Bonferroni corrected and meaningless data are discarded). After masking the important face information, the fit of the neural networks to the psychometric functions, as well as the PSEs, was disturbed to different degrees in both training cases. Specifically, comparing the results for the object and face pre-trained models, the object pre-trained models have a relatively higher bias, i.e. a worse curve fit and more nonsensical PSEs, in all three cases (masking the eyes, nose and mouth), while for the different masking conditions, comparing Fig \ref{psf2} and Fig \ref{psf1}, it is clear that masking the eyes produces the least bias. In particular, for AlexNet, the bias produced is the largest in all conditions. Figure \ref{exp2} shows LayerCAM for different neural networks with different training strategies as well as different masking conditions. In the object pre-training condition, when the eye information is masked, there is some shift in the neural network's attention towards the mouth, and when the nose and mouth information is masked, the neural network still focuses on these regions. In the face pre-training condition, masking the eyes also shows a shift of information from the mouth. 

\begin{table}[t!]
\centering
\caption{Table III represents the PSE statistics of the prediction equations for the four classical neural networks and the modified FE-AlexNet under face pre-training using a prediction equation that masks the key face information isometric variation of the emotionally effective face image input, where the human behavioral experimental data is used as the baseline and the p-values are Bonferroni corrected.}
\begin{tabular}{lllllll}
\hline
                &         &  &  PSE   & \textit{t}(49)   & \textit{p}  \\ \hline
Eyes            & AlexNet &  &  0.707 & 9.696            &  $<$.001    \\
                & VGG11   &  &  0.763 & 12.762           &  $<$.001    \\
                & VGG13   &  &  0.737 & 11.331           &  $<$.001    \\
                & VGG16   &  &  0.715 & 10.116           &  $<$.001    \\
                & FE-AlexNet &  &  0.690 & 8.740         &  $<$.001     \\\hline
Nose            & \textbf{AlexNet} &  &  $\blacktriangle$ & $\blacktriangle$          &  $\blacktriangle$    \\
                & VGG11   &  &  0.585 & 2.912            &  0.070      \\
                & VGG13   &  &  0.760 & 12.575           &  $<$.001    \\
                & VGG16   &  &  0.671 & 7.696            &  $<$.001    \\
                & \textbf{FE-AlexNet} &  &  $\blacktriangle$ & $\blacktriangle$       &  $\blacktriangle$     \\\hline
Mouth           & AlexNet &  &  0.478 & -3.006           &  0.054      \\
                & VGG11   &  &  0.721 & 10.470           &  $<$.001    \\
                & VGG13   &  &  0.664 & 7.293            &  $<$.001    \\
                & VGG16   &  &  0.683 & 8.348            &  $<$.001    \\
                & \textbf{FE-AlexNet} &  &  \textbf{0.235} & \textbf{-16.425}       &  \textbf{$<$.001}     \\\hline
\label{tab3}
\end{tabular}
    \begin{tablenotes}
        \footnotesize
        \item $\blacktriangle$ Indicates that the data had a large bias. thus the results are not meaningful and had been discarded for analysis.
    \end{tablenotes}
\end{table}

\subsubsection{Brief Discussion of Experiment 2}

In this experiment, we further validated using the reverse correlation experimental paradigm. Here, we tested the performance and the processing mechanisms of the CNNs using reverse correlation method: offering them the disrupt visual inputs with essential areas of faces unavailable. The results suggest that different prior knowledge affects the performance of the neural network: the object pre-trained models have a relatively higher bias in subtracting visual inputs for processing, and performed even poorer in terms of the psychometric function.

In addition, comparing the performance of the neural networks under different masking conditions with the same training strategy (i.e., comparing object pre-trained models and face pre-trained models separately), the results favor the notion that the neural networks are more biased towards using information from the lower part of the face in the emotion recognition task, as the effect of masking the eye is minimal. 

More graphically, in LayerCAM results the object-trained neural network is still attempting to recognize information that is not meaningful, while the face recognition network is `correctly' transferring the recognition information to other key information about the face, with the exception of masking the nose and mouth, which has a greater impact on the attentional patterns of the neural network. In addition, masking the nose and mouth had a greater impact on the attentional patterns of the neural network, and combining the results of Experiment 1, we notice that information from the mouth played a more critical role, which is consistent with previous studies \cite{grahlow2022impact}.

\subsection{The Experiment 3: Creating the Feature Excitation AlexNet Based}

\begin{figure*}[t!]
    \centering
    \includegraphics[width=0.75\textwidth]{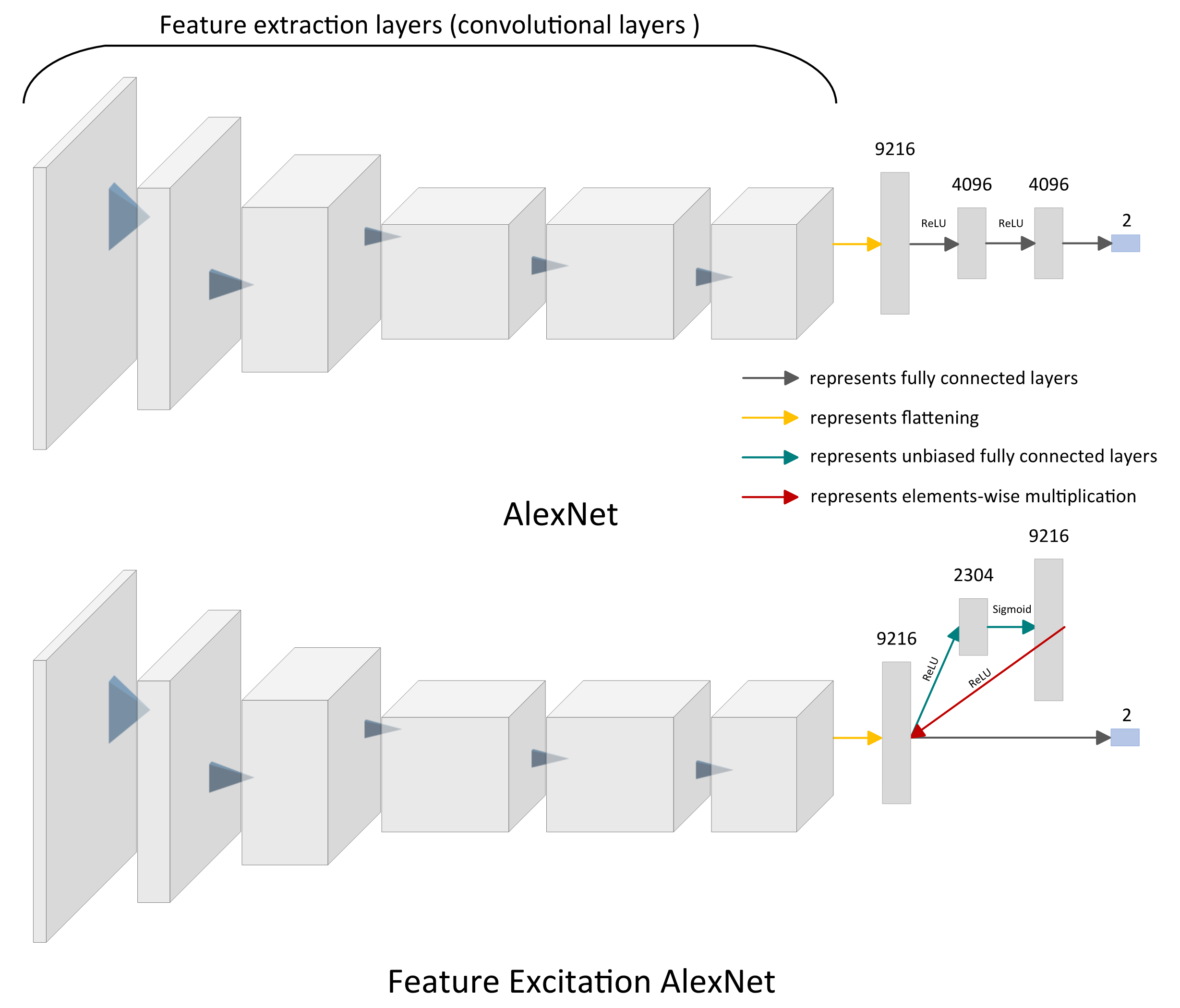}
    \caption{Comparison of Feature Excitation AlexNet and the original AlexNet. We perform an unbiased fully connected auto-correlation operation directly on the feature values after flattening, and then multiply the computed values at the element level with the original feature values using the Sigmoid activation as weights, replacing the original AlexNet fully connected layer with this structure.}
    \label{fealex}
\end{figure*}

The results of the above two experiments show us that models pre-trained on different tasks have different attentional biases that affect the final performance of the model, so here we designed Feature Excitation AlexNet (FE-AlexNet) based on the widely used self-attention module, the Squeeze-and-Excitation module, an algorithm that allows features extracted from the model to be automatically correlated. The FE-AlexNet is shown in Figure \ref{fealex}. We assume that the problem with the model's attention performance is due to an overfitting effect during training, and then the self-attention algorithm can effectively weaken this negative effect. The visual features extract by the FE-AlexNet are the same as the original AlexNet does. Based on the ample evidence from cognitive and neuroscience evidence, it is reasonably assume that the FE-AlexNet can further process the visual inputs and outperform the original AlexNet.

\subsubsection{Results of Experiment 3}

The experimental results for the FE-AlexNet are shown in Figure \ref{psf1} and Figure \ref{psf2} and in Table \ref{tab1}, Table \ref{tab2} and Table \ref{tab3}, shown as black dashed lines in the figures and in bold in the tables. The results in Figure \ref{psf1} and Table \ref{tab1} show that after the introduction of the feature excitation structure replacing the fully connected layer, the psychometric function of the neural network moved closer to human performance compared to the original AlexNet (yellow lines), showing a lower starting point and a non-significant difference in PSE with humans in the object pre-training condition (\textit{t}(49) = 0.501, \textit{p} = 0.875). In reverse correlation experiments, it gives rise to a similar pattern to the original AlexNet.

\subsubsection{Brief Discussion of Experiment 3}

In this part of experiment, we updated the full connected layers of the object pre-trained AlexNet (the one with the worse performance in terms of face perception as well as the oldest structure), based on previous findings in computational neuroscience \cite{chang2017code,chang2021explaining}. By replacing the fully connected layer structure with a feature-activated self-attentive structure, we were able to weaken the perceptual bias induced by pre-train weights of a object pre-traind AlexNet. Moreover, this updated FE-AlexNet outperformed other CNNs with better pre-train weights (i.e., face pre-trained) as well as more complex/human-like structures (i.e., VGGNets). 

The attentional re-calibration, introduced by the feature excitation structure, plays an important role in this update. Previous studies in neuroscience, researchers have suggested that it is possible to decode any face using only low-dimensional IT cortex neurons \cite{chang2017code,chang2021explaining,higgins2021unsupervised}. The operation of fully connected down-sampling in our model has similarities with it. Although FE-AlexNet performs well in conventional tasks, there seems to be room for improvement in the recognition of occluded images with large distortions. Future research will need to introduce additional structures to explore ways to improve its robustness. Nevertheless, our structure successfully exploits the exploration of attentional mechanisms, and the re-calibration of attention and simulation of low-dimensional IT cortex face processing successfully mitigates the bias introduced by the prior knowledge of different neural networks.

\section{Discussion}\label{sec12}

In this study, we tested the transfer learning performance of 4 classic and well-studied (in terms of cognitive science and computer science) CNNs (AlexNet, VGG11, VGG13, VGG16) with different pre-train weights (i.e., object-based vs face-based) on a facial emotion valence categorical test. Hence, we compared their performance against human participants' facial emotion valence perception as a psychophysiology benchmarks. Moreover, based on the findings of this study and previous evidence from cognitive science and neuroscience, we updated the worse-performed object pre-trained AlexNet by replacing the full connection layers with self-attention feature-excitation structures. 

The results of Experiment 1 provide preliminary evidence that different pre-training materials have an impact on neural network performance, and that similar tasks outperform different tasks. The Experiment 2 further validated the results of Experiment 1 using a reverse correlation experimental paradigm, and in addition, we found that the neural network was more biased towards the lower part of the face's important information. This can be explained by an attentional bias, since both the correspondence of the mental functions and the LayerCAM representation point to this result. Therefore, based on this idea, we propose FE-AlexNet, a structure in which we introduce a self-attention mechanism and replace the fully connected layer of the AlexNet network. In Experiment 3, our model succeeded in weakening the bias generated by the neural network in different pre-training contexts. However, the experimental results also show that this model still performs similarly to the original AlexNet in the face of more lossy inputs, which requires future research to further improve this issue.

Our study shows that CNNs have parts that are similar to human performance, such as a preference for information about the mouth during high valence \cite{schurgin2014eye}, and a shift in the areas of interest (AOI) of some CNNs in the face masking condition \cite{bylianto2022face}. However, there are still some dissimilarities, such as CNNs focusing too much on local information, and evidence that CNNs still try to focus on meaningless information in the masking condition. Human face processing has been shown in many studies to be a holistic process \cite{maurer2002many,van2010whole}. Our FE-AlexNet performs fully connected auto-correlation on the convolutional feature processing vectors of the neural network, a process that can be seen as a form of global attention processing, which we believe simulates to some extent the holistic processing of faces by humans. In addition, Previous neuroscience research has argued that it is possible to decode any human face using only low-dimensional IT cortex neurons (e.g. 200) \cite{chang2017code,chang2021explaining,higgins2021unsupervised}. In our model, there is a fully connected downsampling operation for the neural network, and although this downsampling and compression process preserves more than 2,000 neurons in our structure, this downsampling operation also verifies that the neural network can represent faces in the late classification layer with lower dimensionality, and that the structure achieves results more similar to those of humans. Therefore, the simulation of the holistic processing and the low dimensionality of the IT cortex on face processing are potentially important reasons why our model is more human-like.

Marr's theory of visual computation \cite{marr2010vision} suggests that intelligent systems that are not similarly constituted may have similar information processing mechanisms. Numerous recent studies have shown the high feasibility of CNNs inspired by biological visual systems as models for human visual systems\cite{li2022disrupted,lindsay2021convolutional,song2021implementation,zhou2022emerged,nonaka2021brain}. In addition, heuristic adaptations of CNNs using insights from neuroscience or cognitive science are also highly plausible. In our recent study, the late layers of the AlexNet network was shown to be highly correlated with the IT cortex of the human visual system using representational similarity analysis (RSA)\cite{yamins2013hierarchical,yamins2016using,jozwik2018deep,xu2021limits}, and in our study we found considerable differences in the attentional biases of different pre-trained neural networks at the end of the convolutional layer. Based on this finding, we introduced a self-attention module resembling a parallel operation, which was shown to be present throughout the visual pathway. We replaced the original fully connected layer with it, resulting in a more human-like neural network. This new structure successfully mimics the horizontal connections and divisive normalization \cite{lindsay2020attention}, which are present throughout the visual pathway \cite{carandini2012normalization,burg2021learning}, in human IT cortex, and may provide inspiration for future neural network design.

Attention mechanisms encompass several concepts, but here we focus on visual attention. Attention is a complex concept, but we can also define it as when viewing complex scenes, people tend to focus their attention on certain regions and reduce it in other regions in order to acquire and process information more effectively \cite{lindsay2020attention, guo2022attention, niu2021review}. Visual attention is thought to be a cognitive mechanism used to select and enhance information in regions of interest and to reduce or ignore information in irrelevant regions. The attention mechanism of artificial neural networks is of course inspired by the human attention mechanism \cite{lindsay2020attention, guo2022attention, niu2021review}. The process of implementing attention in neural networks often uses weighted allocation to achieve the retention of important information and the attenuation and neglect of irrelevant information. This is exploited in our proposed model, where the information extracted from the convolutional layer of neural network is processed in horizon and normalized to weigh the original features at the elementary level, ignoring unimportant information and redistributing attention to the identified attentional bias problem. The results show that this enhancement successfully uses attention theory to mitigate the effects of attentional bias.

\section{Conclusion}\label{sec13}

In summary, we trained 4 types of neural networks to perform face emotion valence tasks based on object pre-training and face pre-training; using human behavioral data as benchmark, disruptive input for reverse correlation analysis, and LayerCAM to indicate CNNs' attention; and also proposed a new neural network structure using self-attention. The main findings of our study are that 

\begin{enumerate}[(1)]

  \item CNNs have similar but not the same visual computation mechanisms as humans in face perception

  \item the CNNs with different pre-train weights and architectures performed differently because of their attention and information extraction mechanisms

  \item the new feature excitation structure with self-attention mechanisms would outperform other classical CNNs

  \item this study offers a new paradigm for future research using human cognitive-behavioral data to improve deep convolutional neural networks

\end{enumerate}

\backmatter

\section*{Data availability}

The code for neural network training and testing can be found in Open Science Framework (https://osf.io/53zpg/?view\_only=e96b62d6ed26442db52864e5236fbbeb). Further information and requests for resources and reagents should be directed to and will be fulfilled by the lead contact, Yiyang Chen (yychen90@suda.edu.cn).

\section*{Acknowledgments}

This work was supported in part by National Natural Science Foundation of China under Grant 62103293 and Grant 32200850, in part by Natural Science Foundation of Jiangsu Province under Grant BK20210709 and Grant BK20200867, in part by Suzhou Municipal Science and Technology Bureau under Grant SYG202138, and in part by Entrepreneurship and Innovation Plan of Jiangsu Province under Grant JSSCBS202030767 and Grant JSSCBS20210641. 

\section*{Compliance with Ethical Standards}

\subsection*{Disclosure of potential conflicts of interest}

The authors declare no conflict of interest.

\subsection*{Ethical Approval}

This study was approved by the Ethics Committee of Soochow University. 

\subsection*{Informed consent}

All individual participants offered written consent before the experiment. 

\bibliography{sn-bibliography}

\end{document}